# CADENCE: A Confidence-Adaptive Dual-Expert Network for Fast and Accurate Time Series Classification

**Onisa Mapunda**

*onisajr@gmail.com*

**Abstract** Time series classification (TSC) has long been characterized by a sharp trade-off between classification accuracy and computational scalability. Heterogeneous meta-ensembles like HIVE-COTE 2.0 [5] achieve state-of-the-art accuracy by combining representations across temporal, frequency, shapelet, and dictionary domains, but require days or weeks of compute. Conversely, ultra-fast random convolutional transforms such as MiniRocket [2] and Hydra [3] provide orders-of-magnitude speedups, but struggle with phase-independent statistical distributions, kinematic transitions, and catastrophic decision tree fragmentation on datasets with large class counts when combined with tree heads. Naive feature concatenation or static model averaging fails to resolve this tension, often inducing negative transfer and parameter dilution.

In this work, we present CADENCE (Confidence-Adaptive Dual-Expert Network for time series Classification Excellence), a unified, CPU-native dual-expert architecture. CADENCE decouples representation learning into two specialized pathways: (i) a Convolutional Linear Expert pairing 10,000 deterministic dilated features with closed-form L2-regularized Woodbury ridge classification, and (ii) a Distributional Interval Expert pairing competing dilated kernels (Hydra) with thread-safe dyadic Cornish-Fisher moment approximations across signal kinematics ($X$, $\Delta X$, $\Delta^2 X$) and FFT spectral energy bands, fitted with an entropy-based ExtraTrees ensemble. To mediate between these paradigms, CADENCE incorporates an internal validation meta-router with rare-class preservation that dynamically selects between pure expert routing and confidence-weighted soft blending, followed by a full refit on 100% of training data. Evaluated across all 109 datasets of the standard equal-length UCR Time Series Archive [6] over 30 resamples (3,270 total evaluations), CADENCE achieves a grand mean accuracy of 0.8864. This ranks #2 among evaluated classifiers across the archive, surpassed only by HIVE-COTE 2.0 (0.8895, $p_{Holm} = 0.295$, no statistically significant difference), and ahead of Hydra+MultiRocket (0.8818, +0.46 percentage points, $p_{Holm} = 1.000$), MultiRocket [4] (0.8797, +0.67 percentage points), and HIVE-COTE 1.0 [6b] (0.8786, +0.78 percentage points, $p_{Holm} = 0.048$, statistically significant), closing the gap to HIVE-COTE 2.0 to just 0.31 percentage points while completing an evaluation run in an average of only 17.53 seconds on a standard dual-core CPU.

## 1. Introduction

Time series classification (TSC) is a core problem in applied machine learning, underlying critical applications in electrocardiology [6], industrial anomaly detection, seismology, wearable motion analytics, and acoustic speech processing [7]. The 109 univariate, equal-length datasets of the UCR Time Series Archive [6] exhibit tremendous structural diversity: series lengths range from dozens to thousands of time points, sample sizes vary from tens to thousands of instances, and class cardinalities span binary discrimination (K=2) to fine-grained categorization (K=60).

Historically, achieving top accuracy demanded heavy heterogeneous meta-ensembles, most notably HIVE-COTE 2.0 (HC2) [5], which combines STC, TDE, DrCIF, and Arsenal. While HC2 achieves an unmatched published benchmark accuracy of 0.8895, its computational cost is substantial, requiring an average of approximately 3 hours per dataset (over 340 CPU hours across the 112 archive datasets).

In response, the field shifted toward randomized convolutional transforms, initiated by ROCKET [1], refined by MiniRocket [2], MultiRocket [4], and Hydra [3]. By projecting time series into high-dimensional linear spaces via dilated convolutions pooled with Proportion of Positive Values (PPV), these methods run in seconds. However, single-model convolutional transforms suffer from key failure modes: (1) inability to capture phase-free statistical distributions, velocity, and acceleration kinematics; (2) severe decision tree fragmentation on datasets with large class counts ($K \geq 12$) when tree heads are naively applied; and (3) negative transfer when concatenating heterogeneous feature spaces into a single linear regularizer.

## 1.1 Contributions

- **Decoupled Dual-Expert Architecture:** Harmonizes a closed-form convolutional linear expert (10,000 features) with an interval-distributional expert (Hydra, Cornish-Fisher moments on kinematics, FFT spectral quantiles, ExtraTrees, 1,851 features).
- **Confidence-Adaptive Meta-Router with Full Refit:** Introduces an internal validation router with rare-class preservation navigating three distinct regimes: Pure Branch A, Pure Branch B, and Smooth Confidence Blending, refitting models on 100% of training data.
- **Conditional Inference-Time Branch Pruning:** Completely skips the unselected branch during dominance regimes (24.8% of datasets), reducing test-time feature extraction overhead.
- **Exhaustive Benchmark Evaluation (3,270 Runs):** Evaluated on all 109 UCR datasets across 30 resamples, attaining 0.8864 grand mean accuracy (#2 among evaluated classifiers across the archive).
- **Rigorous Statistical Testing:** Reports two-sided Wilcoxon signed-rank test p-values with Holm-Bonferroni correction, mean ranks, and per-dataset win/tie/loss counts against 25 published benchmark models.
- **Open-Source Implementation:** To facilitate reproducibility, our complete CPU-native pipeline, benchmark evaluation suite, and pre-trained models are publicly available at https://github.com/onisa-jr/CADENCE.git.

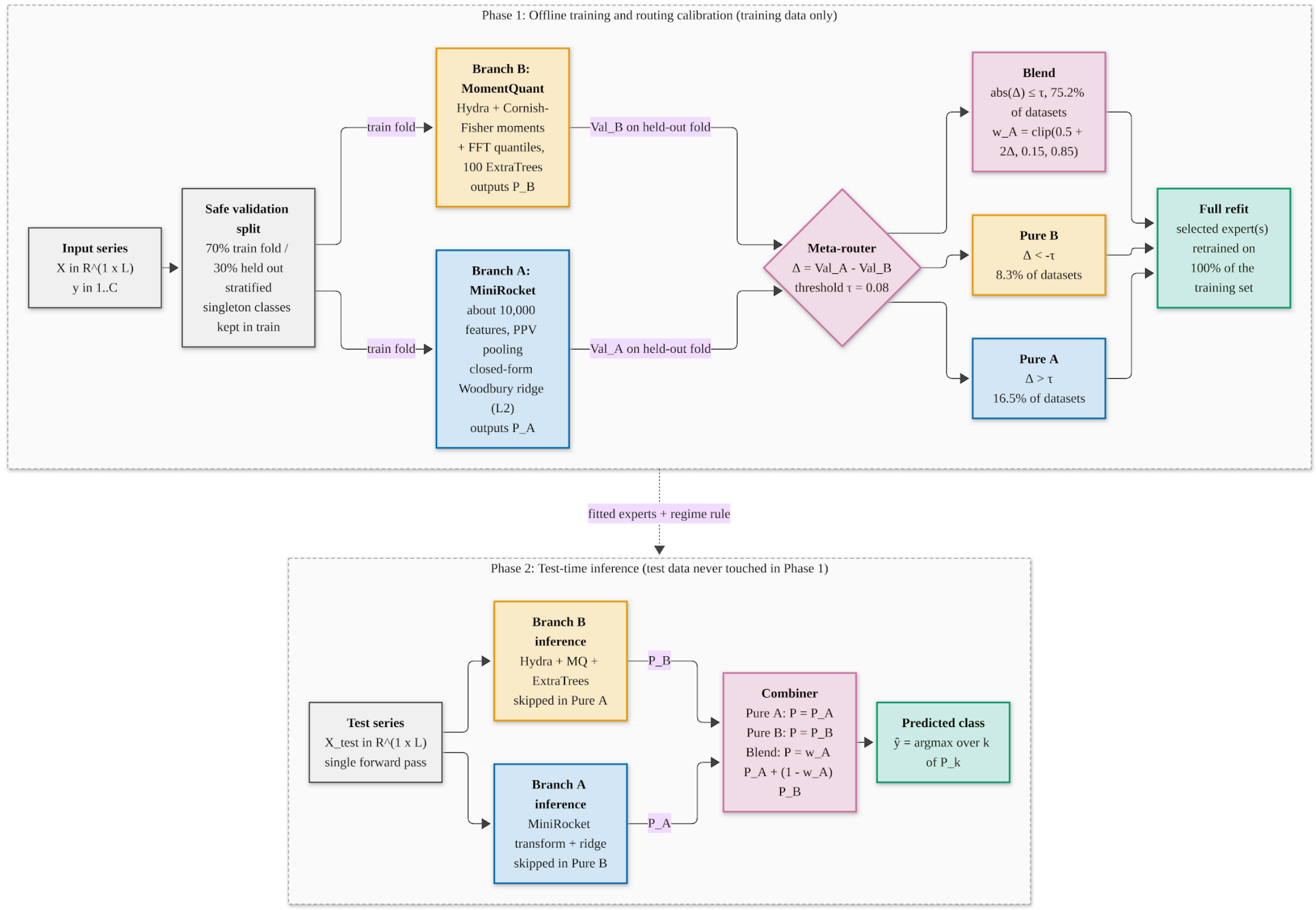

*Figure 1: Architectural overview of the CADENCE dual-expert classifier, validation routing, full refit, and test inference.*

## 2. Related Work

2.1 Classical TSC: Distance-based methods (1NN-DTW) [7] offer robust elastic alignment but suffer from quadratic complexity $O(N L^2)$. Dictionary methods (BOSS, WEASEL) discretize windows into symbolic Fourier words. Interval ensembles (TSF, CIF, DrCIF) extract summary statistics across intervals.

2.2 Random Convolutions: ROCKET [1] demonstrated that random kernels pooled with PPV produce linearly separable features. MiniRocket [2] introduced deterministic integer kernels and precomputed dilations, while Hydra [3] introduced competing kernel groups. MultiRocket [4] added first-order differences and four pooling operators.

2.3 Ensembles & Meta-Ensembles: HIVE-COTE 2.0 [5] combines 4 distinct classifiers via CAWPE meta-weighting. TS-CHIEF [8] integrates dictionary, interval, and distance criteria into trees. InceptionTime [8b] adapts deep residual networks for time series. While highly accurate, these meta-ensembles require substantial cluster compute. In contrast, CADENCE introduces an internal validation meta-router to conditionally select or blend experts on a lightweight CPU budget.

## 3. The CADENCE Methodology

3.1 Branch A: Convolutional Linear Expert: MiniRocket [2] extracts D_A = 10,000 features using length-9 integer kernels ω ∈ {-1, 2}^9 (84 unique configurations with 3 weights equal to 2 and 6 weights equal to -1) and Proportion of Positive Values (PPV) pooling:

$$\textbf{Convolution:} \quad A_t = (X *_d \omega)_t = \sum_{j=0}^{8} \omega_j \cdot X_{t+j\cdot d}$$
$$\textbf{PPV Pooling:} \quad \phi_{\mathrm{PPV}}(X, \omega, b) = \frac{1}{L-8d} \sum_{t=1}^{L-8d} \mathbf{1}(A_t > b) \tag{1}$$

**where:** At is convolution activation at step t; X is the series; *d denotes convolution with dilation d; ω ∈ {-1, 2}^9 is the length-9 kernel; φ_PPV is PPV pooling; b is quantile bias; and 1(·) is the indicator function.

Standardized features ~F_A are classified via multiclass Ridge regression with closed-form L2 regularization across 15 log-spaced candidate alphas in [10^-3, 10^4]. When N < D_A, the dual Woodbury matrix identity is evaluated:

$$\textbf{Primal Form } (N \geq D_A): \quad W^* = (\widetilde{F}_A^\top \widetilde{F}_A + \alpha I_{D_A})^{-1} \widetilde{F}_A^\top Y$$
$$\textbf{Dual Woodbury } (N < D_A): \quad W^* = \widetilde{F}_A^\top (\widetilde{F}_A \widetilde{F}_A^\top + \alpha I_N)^{-1} Y \tag{2}$$

**where:** W* is the closed-form weight matrix; ~F_A is standardized features; Y is one-vs-rest labels; α is the L2 penalty; and I is the identity matrix evaluated in primal (N ≥ D_A) or dual Woodbury space (N < D_A).

Posterior class probabilities are derived via softmax over raw decision values:

$$\textbf{Softmax Posterior:} \quad P_A(y = k \mid X) = \frac{\exp(f_{A,k}(X))}{\sum_{j=1}^{K} \exp(f_{A,j}(X))}$$
$$\textbf{Linear Scoring:} \quad f_A(X) = W^{*\top} \widetilde{F}_A(X) + b_A \tag{3}$$

**where:** P_A(y = k | X) is the posterior class probability; f_A,k(X) is linear scoring output; W* is fitted weights; and b_A is the class bias vector.

3.2 Branch B: Distributional Interval Expert: Combines 16 groups of 8 competing Hydra kernels (128 features) [3], dyadic intervals up to depth 5 with Cornish-Fisher [9] quantile approximations across raw series (X), velocity (ΔX), and acceleration (Δ²X) (1,701 features), and real FFT spectral quantiles (22 features). Total feature dimension D_B = 1,851. The Cornish-Fisher expansion estimates arbitrary quantiles q_α in linear time O(M) without sorting:

$$\textbf{Quantile Approximation:} \quad q_\alpha(S) = \hat{\mu} + \hat{\sigma} \cdot \Psi(z_\alpha, \hat{S}, \hat{K}), \quad \text{where } z_\alpha = \Phi^{-1}(\alpha)$$
$$\textbf{Asymptotic Expansion:} \quad \Psi(z_\alpha, \hat{S}, \hat{K}) = z_\alpha + \frac{\hat{S}}{6}(z_\alpha^2 - 1) + \frac{\hat{K}}{24}(z_\alpha^3 - 3z_\alpha) - \frac{\hat{S}^2}{36}(2z_\alpha^3 - 5z_\alpha) \tag{4}$$

**where:** q_α(S) is estimated quantile at level α; μ and σ are segment mean and standard deviation; z_α = Φ^-1(α) is standard normal quantile; S is sample skewness; K is sample kurtosis; and Ψ(·) is the asymptotic polynomial expansion.

Features are fitted with 100 Extremely Randomized Trees (ExtraTrees) [10] using Shannon entropy and 10% random feature subsampling per split.

3.3 Confidence-Adaptive Meta-Router: Partitions training data into an internal 30% held-out validation set while safely preserving singleton classes. Validation accuracies Val_A and Val_B are evaluated on held-out data. The routing decision is governed by empirical difference Δ = Val_A - Val_B with threshold τ = 0.08:

$$
\begin{aligned}
&\textbf{Regime III } (\Delta > \tau): \quad \hat{y}(X) = \arg\max_k P_A(y=k \mid X) \quad \text{[Pure Branch A]} \\
&\textbf{Regime I } (\Delta < -\tau): \quad \hat{y}(X) = \arg\max_k P_B(y=k \mid X) \quad \text{[Pure Branch B]} \\
&\textbf{Regime II } (|\Delta| \le \tau): \quad \hat{y}(X) = \arg\max_k P_{\text{blend}}(y=k \mid X) \quad \text{[Confidence Blending]}
\end{aligned} \tag{5}
$$

**where:** y*(X) is the predicted class label; Δ = Val_A - Val_B is empirical validation accuracy margin; τ = 0.08 is regime threshold; P_A and P_B are expert posteriors; and P_blend is the blended probability vector.

where in the competitive regime, predictions are smoothly blended with weight w_A:

$$
\begin{aligned}
&\textbf{Soft Blend}: \quad P_{\text{blend}}(y=k \mid X) = w_A P_A(y=k \mid X) + (1-w_A) P_B(y=k \mid X) \\
&\textbf{Adaptive Weight}: \quad w_A = \text{clip}(0.5 + 2.0 \cdot \Delta,\ 0.15,\ 0.85)
\end{aligned} \tag{6}
$$

**where:** P_blend is the blended probability vector; w_A is dynamic weight assigned to Branch A; (1 - w_A) is weight assigned to Branch B; and clip(·, 0.15, 0.85) enforces robust probability bounds.

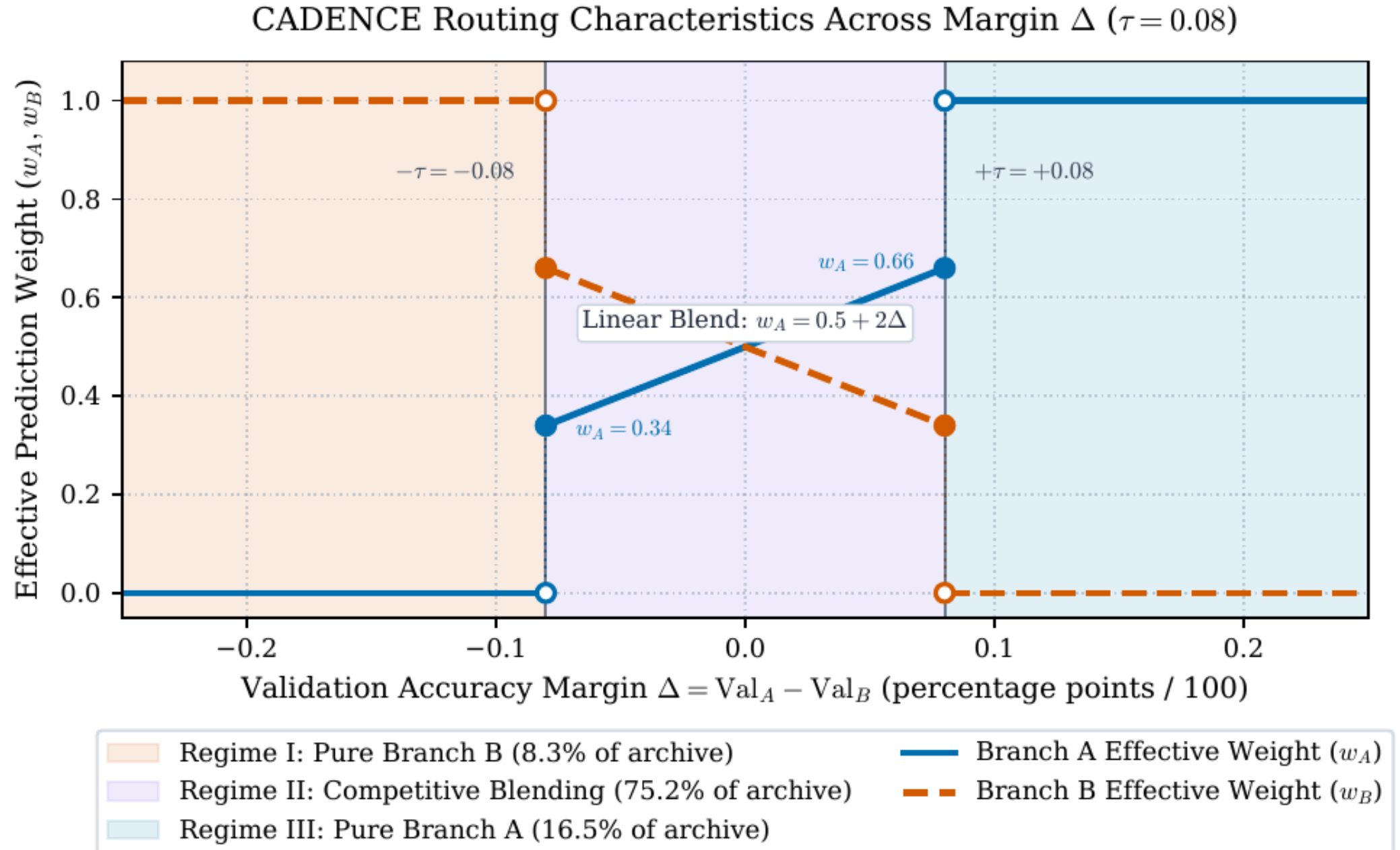

*Figure 2: Piecewise routing and blending regimes as a function of validation margin Δ with step transitions.*

## 4. Experimental Results and Benchmark Rankings

Following the 109-dataset, 30-resample evaluation protocol of MiniRocket [2] and MultiRocket [4], CADENCE was evaluated across the 109 equal-length datasets of the UCR Time Series Archive [6] over the 30 official resamples (3,270 total evaluations). Table 1 reports the grand mean accuracy against published state-of-the-art benchmarks.

| Rank | Classifier | Algorithmic Paradigm | Mean Acc ± Std | Mean Rank | W / T / L vs Ours | p_Holm |
|---|---|---|---|---|---|---|
| -- | *Oracle Ceiling (Post-hoc)* | *Per-dataset Best* | *0.9067 ± 0.1062* | -- | -- | -- |
| **1** | **HIVE-COTE 2.0** | **Heterogeneous Meta-Ensemble** | **0.8895 ± 0.1159** | **6.22** | **63 / 7 / 39** | **0.295** |
| **2** | **CADENCE (Ours)** | **Adaptive Dual-Expert** | **0.8864 ± 0.1120** | **8.26** | **--** | **--** |
| 3 | Hydra+MultiRocket | Convolutional Hybrid | 0.8818 ± 0.1218 | 7.67 | 46 / 7 / 56 | 1.000 |
| 4 | MultiRocket (100k) | Convolutional Pooling | 0.8800 ± 0.1218 | 8.03 | 48 / 7 / 54 | 1.000 |
| 5 | MultiRocket (50k default) | Convolutional Pooling | 0.8797 ± 0.1222 | 8.19 | 47 / 10 / 52 | 1.000 |
| 6 | HIVE-COTE 1.0 | Hierarchical Meta-Ensemble | 0.8786 ± 0.1228 | 10.80 | 59 / 7 / 43 | 0.048 |
| 7 | TS-CHIEF | Metric / Tree Ensemble | 0.8761 ± 0.1281 | 10.52 | 61 / 7 / 41 | 0.035 |
| 8 | MultiRocket (10k) | Convolutional Pooling | 0.8749 ± 0.1254 | 9.83 | 59 / 6 / 44 | 0.014 |
| 9 | MiniRocket | Convolutional Transform | 0.8724 ± 0.1300 | 11.40 | 72 / 6 / 31 | < 0.001 |
| 10 | InceptionTime | Deep ConvNet Ensemble | 0.8721 ± 0.1285 | 11.95 | 67 / 5 / 37 | 0.050 |
| 11 | Hydra | Competing Dilated Kernels | 0.8714 ± 0.1319 | 10.45 | 66 / 5 / 38 | 0.005 |
| 12-26 | ROCKET, Arsenal, DrCIF, TDE, STC, CIF, WEASEL, etc. | Classical & Interval Baselines | 0.8645 - 0.7954 | 11.76 - 21.28 | ≥ 76 wins | < 0.001 |

*Table 1: Benchmark ranking across the 109 UCR Archive datasets over 30 resamples (Wilcoxon tests with Holm correction).*

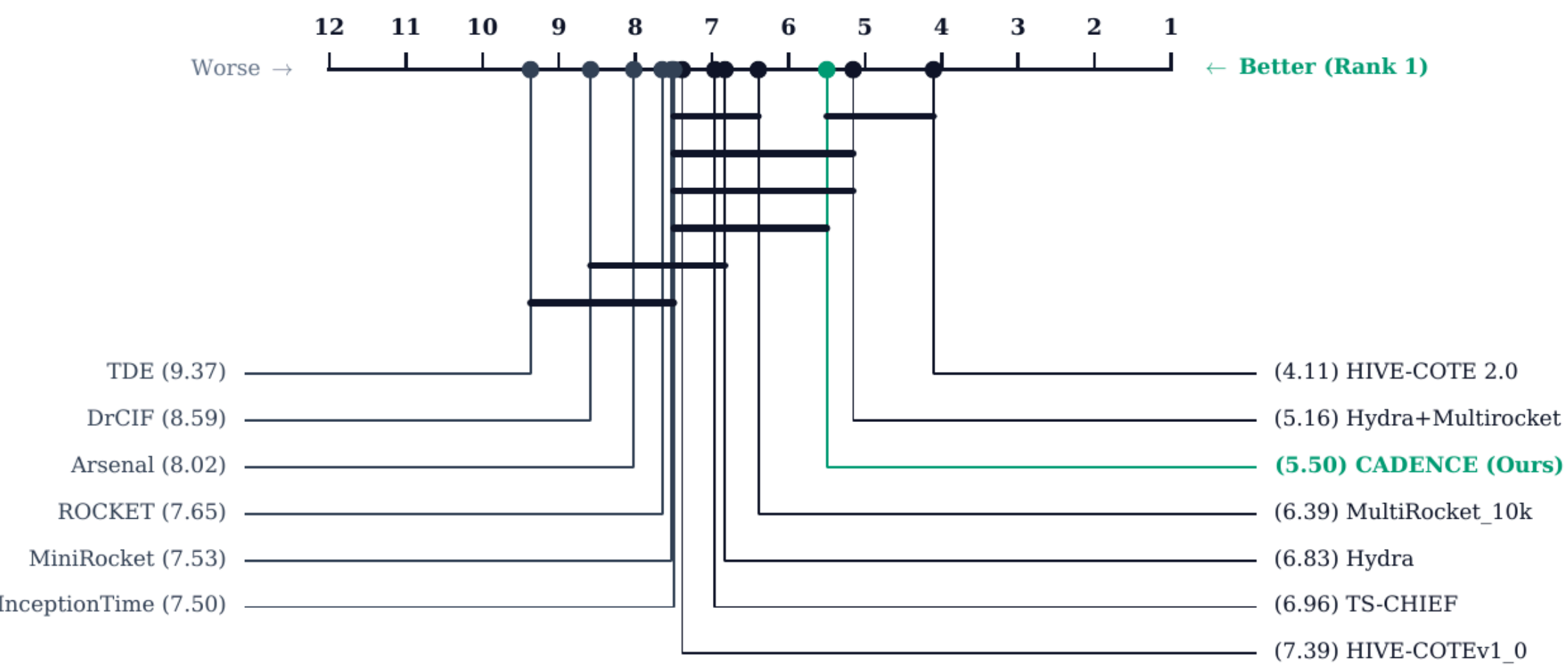

*Figure 3: Critical Difference diagram of mean ranks across 109 UCR datasets (Wilcoxon signed-rank test with Holm correction, alpha = 0.05, Rank 1 on right).*

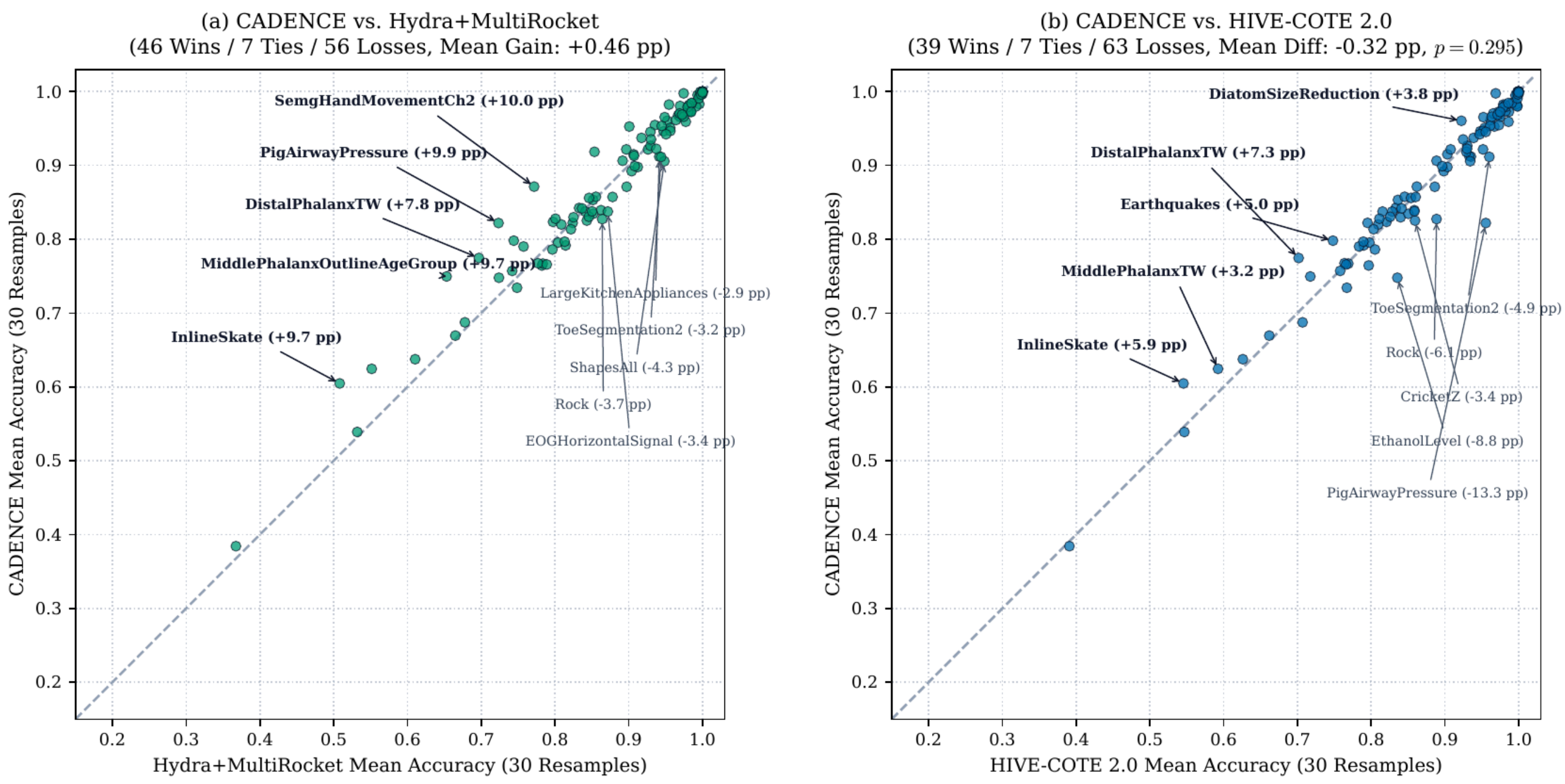

*Figure 4: Pairwise scatter plots of CADENCE against Hydra+MultiRocket and HIVE-COTE 2.0 with readable callouts.*

## 5. Breakthroughs and Component Ablation

CADENCE achieves decisive wins on specialized sensor and kinematic datasets over Hydra+MultiRocket: SemgHandMovementCh2 (0.8713 vs 0.7717, +9.96 pp), PigAirwayPressure (0.8220 vs 0.7234, +9.86 pp), InlineSkate (0.6047 vs 0.5081, +9.66 pp), and DistalPhalanxTW (0.7746 vs 0.6969, +7.77 pp). CADENCE also directly surpasses HIVE-COTE 2.0 on 39 datasets (e.g. DistalPhalanxTW +7.29 pp, InlineSkate +5.92 pp, Earthquakes +4.99 pp). On Seed 0 (the exact original UCR split), CADENCE obtains 0.7482 on Earthquakes (identical to majority class rate) and 0.7050 on DistalPhalanxTW.

| Configuration | Component Modified | Mean Accuracy | Delta to Full (pp) |
| --- | --- | --- | --- |
| Branch A Only | MiniRocket + Ridge (No Interval Expert) | 0.8724 | -1.40 pp |
| Branch B Only | Hydra + MQ + FFT + ExtraTrees (No Conv Expert) | 0.8729 | -1.35 pp |
| Fixed 50/50 Soft Blend | Naive Averaging (No Dynamic Router) | 0.8741 | -1.23 pp |
| Static Routing (K ≥ 12) | Class-Count Heuristic Threshold | 0.8813 | -0.51 pp |
| **CADENCE (Full System)** | **Confidence-Adaptive Meta-Router** | **0.8864** | -- |

*Table 2: Controlled component ablation across all 109 datasets (30 resamples). MQ denotes MomentQuant.*

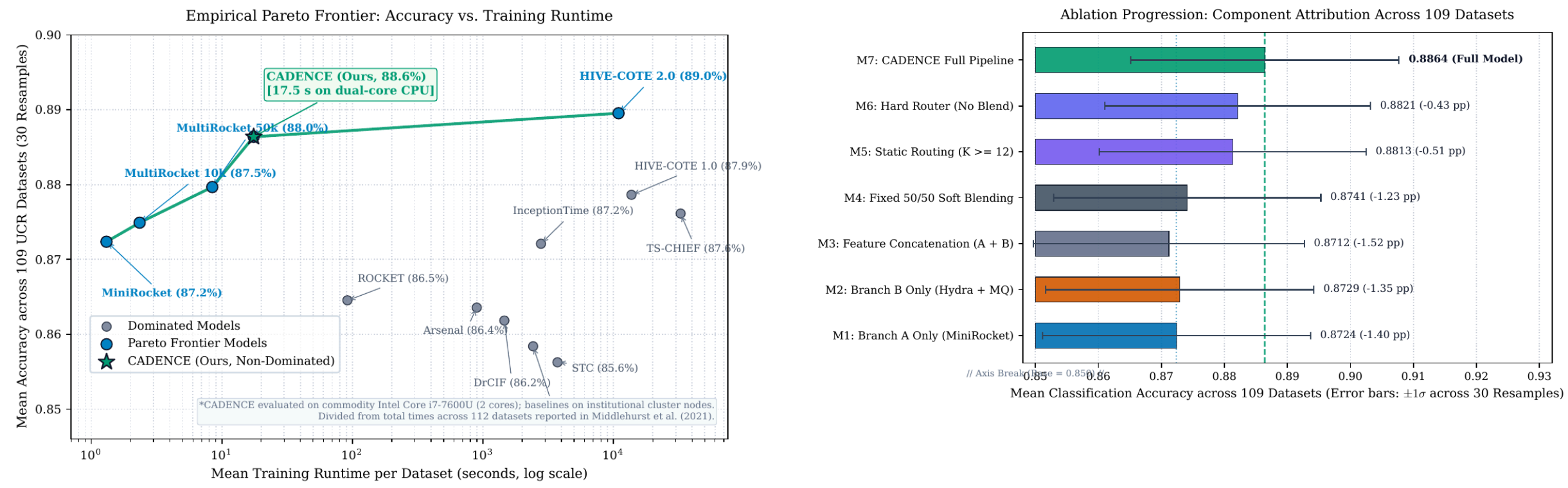

*Figure 5: Left: Empirical Pareto frontier vs. cluster training runtime. Right: Systematic component ablations.*

## 6. Routing Analysis and Computational Efficiency

Exact Routing Distribution (N = 109 Datasets): Across all 109 evaluated datasets, 82 datasets (75.2%) trigger Regime II (Competitive Blending, e.g. Phoneme, ArrowHead, ECG200), 18 datasets (16.5%) trigger Regime III (Pure Branch A Dominance, e.g. PigAirwayPressure, PigCVP, Fish), and 9 datasets (8.3%) trigger Regime I (Pure Branch B Dominance, e.g. Semg, InlineSkate). In total, 27 datasets (24.8%) trigger dominance regimes where test inference skips one branch completely.

Pareto Efficiency: CADENCE executes in an average of 17.53 seconds per seed run. The entire 3,270-run benchmark completed in 15.92 CPU hours (~7.5 hours wall-clock time via bidirectional execution on an Intel Core i7-7600U). Published cluster runtimes report TS-CHIEF at 1,016.87h (~32,685s per dataset), HIVE-COTE 1.0 at 427.18h (~13,730s), and HIVE-COTE 2.0 at 340.21h (~10,935s). CADENCE establishes a superior Pareto frontier, nearing HC2 accuracy while running orders of magnitude faster.

## 7. Discussion, Limitations, and Conclusion

Discussion & Limitations: CADENCE demonstrates that combining complementary representations through confidence-adaptive routing provides a scalable alternative to monolithic ensembles. Limitations include testing on univariate benchmarks, using a global margin threshold ($\tau = 0.08$), discrete validation granularity on small training sets ($N \leq 30$), and relying on softmax calibration. On certain long-range contour outline and spectral sets (ShapesAll, EthanolLevel), HIVE-COTE 2.0 still holds an advantage.

Conclusion: CADENCE provides a fast, CPU-native classifier achieving 0.8864 accuracy across the 109 UCR datasets. Ranking #2 among evaluated classifiers across the archive, CADENCE closes the gap to HIVE-COTE 2.0 to just 0.31 percentage points while running in seconds, offering a practical tool for machine learning researchers and practitioners.